\documentclass[acmtog,natbib=true]{acmart}
\acmSubmissionID{271}

\usepackage{booktabs} 

\usepackage{bm}
\usepackage{cleveref}
\crefname{section}{§}{§§}
\Crefname{section}{§}{§§}
\usepackage{enumitem}

\usepackage[ruled]{algorithm2e}
\SetKwInput{KwInput}{Input}                
\SetKwInput{KwReturn}{Return}
\SetKwInput{KwIntialize}{Initialize} 

\SetAlFnt{\small}
\SetAlCapFnt{\small}
\SetAlCapNameFnt{\small}
\SetAlCapHSkip{0pt}

\newcommand\blfootnote[1]{%
  \begingroup
  \renewcommand\thefootnote{}\footnote{#1}%
  \addtocounter{footnote}{-1}%
  \endgroup
}

\acmJournal{TOG}
\acmVolume{40}
\acmNumber{6}
\acmArticle{219}
\acmYear{2021}
\acmMonth{12}

\setcopyright{rightsretained}

\acmDOI{10.1145/3478513.3480528}

\begin{document}
\title{Neural Actor: Neural Free-view Synthesis of Human Actors with Pose Control}

\author{Lingjie Liu}
\affiliation{%
	\institution{Max Planck Institute for Informatics}
	\country{Germany}
}
\email{lliu@mpi-inf.mpg.de}

\author{Marc Habermann}
\affiliation{%
	\institution{Max Planck Institute for Informatics}
	\country{Germany}
}
\email{mhaberma@mpi-inf.mpg.de}

\author{Viktor Rudnev}
\affiliation{%
	\institution{Max Planck Institute for Informatics}
	\country{Germany}
}
\email{vrudnev@mpi-inf.mpg.de}

\author{Kripasindhu Sarkar}
\affiliation{%
	\institution{Max Planck Institute for Informatics}
	\country{Germany}
}
\email{ksarkar@mpi-inf.mpg.de}

\author{Jiatao Gu}
\affiliation{\institution{Facebook AI Research}
\country{USA}
 }
 \email{jgu@fb.com}
 
\author{Christian Theobalt}
\affiliation{\institution{Max Planck Institute for Informatics}
\country{Germany}
 }
 \email{theobalt@mpi-inf.mpg.de}

\newcommand{\revise}[1]{{\textcolor{black}{#1}}}
\newcommand{\reviset}[1]{{\textcolor{blue}{#1}}}


\begin{abstract}

We propose Neural Actor (NA), a new method for high-quality synthesis of humans from arbitrary viewpoints and under arbitrary controllable poses. 
Our method is developed upon recent neural scene representation and rendering works which learn  representations of geometry and appearance from only 2D images. While existing works demonstrated compelling  rendering of static scenes and playback of dynamic scenes, photo-realistic reconstruction and rendering of humans with neural implicit methods, in particular under user-controlled novel poses, is still difficult. To address this problem, we utilize a coarse body model as a  proxy to unwarp the surrounding 3D space into a canonical pose. A neural radiance field learns pose-dependent geometric deformations and pose- and view-dependent appearance effects in the canonical space from multi-view video input. To synthesize novel views of high-fidelity dynamic geometry and appearance, NA leverages 2D texture maps defined on the body model as latent variables for predicting residual deformations and the dynamic appearance. Experiments demonstrate that our method achieves better quality than the state-of-the-arts on playback as well as novel pose synthesis, and can even generalize well to new poses that starkly differ from the training poses. 
Furthermore, our method also supports shape control on the free-view synthesis of human actors. 
\end{abstract}

%
%


%
%


\begin{teaserfigure}
  \centering
    \includegraphics[width=\textwidth]{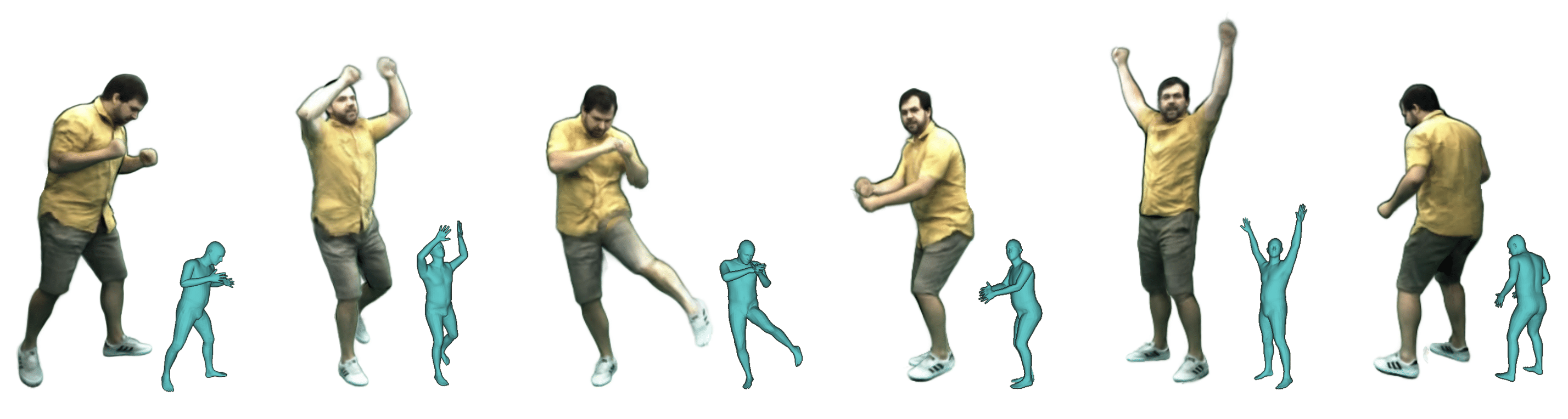} 
    \captionof{figure}{
    Novel view synthesis of an actor using \emph{Neural Actor (NA)} under the control of novel poses,  with the corresponding template mesh models shown at lower right. All the poses are randomly sampled from the testing sequence. 
    } 
    \label{fig:teaser} 
\end{teaserfigure}

\maketitle

\blfootnote{Project page: \href{https://vcai.mpi-inf.mpg.de/projects/NeuralActor/}{\color{magenta}{\url{https://vcai.mpi-inf.mpg.de/projects/NeuralActor/}}}}

%
\section{Introduction}
%
%
Traditional methods for free-viewpoint video generation of humans from multi-view video input employed passive photogrammetric methods or template fitting approaches to capture explicit models of the dynamic geometry and appearance of the moving human~\cite{Xu:SIGGRPAH:2011,Li:2017,Casas:2014,Volino2014,Carranza:2003, borshukov2005universal,Li:2014, zitnick2004high,collet2015high}. 
Novel views are synthesized with classical graphics renderers. 
Capturing such explicit moving human models from images is a very complex, time-consuming and potentially brittle process. 
It is therefore hard to achieve photo-realistic free-viewpoint video quality for humans in diverse apparel. 
Furthermore, these techniques require animatable person-specific surface templates which need sophisticated reconstruction and rigging techniques for creating them.
%
%
\par
Recently neural scene representation and  rendering methods~\cite{tewari2020state} have been presented to overcome many limitations of the aforementioned earlier approaches based on explicit computer graphics modeling and rendering techniques.
They implicitly learn representations of shape and appearance from images, which can be rendered from new viewpoints without requiring explicit computer graphics models.
However, while these approaches show compelling results on static scenes, applying them to high-quality free-viewpoint rendering of humans in general clothing, let alone under novel user-controlled poses, is still difficult.
%
\par
In this paper, we present a new approach, {\em Neural Actor (NA)}, for high-quality free-viewpoint rendering of human actors in everyday attire. NA can play back captured motion sequences with a large number of poses as well as  synthesize free-viewpoint animations under user-controlled novel pose sequences.
NA takes as input multi-view images of a human actor as well as the tracked poses of the actor on the basis of a coarse parametric shape model (SMPL)~\cite{SMPL:2015}.
One challenge we need to address is that simply extending existing neural representations with a pose vector conditioning is not enough (see \emph{NeRF+pose} in  Figure~\ref{Qualitative_novel_pose_compare}) to achieve high-quality pose-dependent renderings.
%
%
Instead, we explicitly deform the posed space to the canonical pose space with an inverse skinning transformation using the SMPL model~\cite{Huang:ARCH:2020}.
We then predict residual deformation~\cite{tretschk2020nonrigid} for each pose with a deformation network, followed by learning pose-conditioned neural radiance fields in the canonical space. 
This design enables us to efficiently handle large movements.
%
\par
However, the above formulation can still lead to blurry rendering results (see \emph{NA w/o texture} in Figure~\ref{ablation1}). 
This is due to the complex dynamics of the surface, pose tracking errors, and the fact that because of various dynamic effects the mapping from the skeletal pose to dynamic geometry and appearance is not a bijection, which therefore cannot be learned reliably using a deterministic neural model with maximum likelihood objectives (e.g. L1/L2 loss), such as NeRF~\cite{mildenhall2020nerf}.
Hence, in order to better capture pose-dependent local shape and appearance changes, we incorporate 2D texture maps defined on the SMPL model as latent variables into the scene representation and break down the mapping into two parts, one from the skeletal pose to pose-dependent texture map and the other from the  pose-dependent texture map to dynamic effects. The uncertainty in the mapping from the skeletal pose to dynamic effects can be mitigated in the former part with the adversarial loss to prevent the training from converging to the mean appearance.

In summary, our contributions are: 
%
\begin{itemize}
    \item We propose \emph{Neural Actor (NA)}, a new method for realistic free-view synthesis of moving human actors with dynamic geometry and appearance. It can play back long pose sequences and synthesize results for challenging new user-controlled poses. 
    \item We present a new strategy to learn a deformable radiance field with SMPL as guidance. It disentangles body movements into inverse skinning transformations and dynamic residual deformations where only the latter needs to be learned.
    \item NA achieves high-quality dynamic geometry and appearance prediction without blurry artifacts by incorporating 2D texture maps defined on SMPL as latent variables.
    \item We captured a new multi-view human performance dataset with dense camera arrays, which contains four sequences of human actors performing various motions. We will make this dataset publicly available. 
\end{itemize}

\textbf{Scope.} 
Since our method leverages a SMPL model for unwarping to the canonical space, it can only handle those clothing types that follow the topological structure of the SMPL model. We regard the issue of handling loose cloth, such as skirts, as future work and discuss it in Section~\ref{sec.limitation}.

%
%
\section{Related Work}
We will review learning-based approaches to neural scene representation and rendering, and generative models for humans.
There are earlier non-learning-based works for video-based character creation~\cite{Xu:SIGGRPAH:2011,Li:2017,Casas:2014,Volino2014} and free-viewpoint videos~\cite{Carranza:2003, borshukov2005universal, Li:2014, zitnick2004high, collet2015high}, which we omit as they are conceptually less related.
%

\paragraph{Neural Scene Representation and Rendering.}
Neural scene representation and rendering algorithms aim at learning scene representations for novel view synthesis from only 2D images. 
Related works in this area can be categorized into the rendering of static and dynamic scenes.
%
Generative Query Networks (GQN)~\citep{eslami2018neural}  represent a 3D scene as a vectorized embedding and use the embedding to render novel views. However, since they do not learn geometric scene structure explicitly, their renderings are rather coarse. 
\citet{DIBR19} propose an approach to differentiable rasterization by applying local interpolation and distance-based global aggregation for foreground and background pixels, respectively.
DeepVoxels~\cite{Sitzmann2019DV} represents a static scene as voxel grids to where learnable features are attached. 
SRN~\cite{sitzmann2019scene} replaces the discretized representation with a continuous learnable function.
Recently, NeRF~\cite{mildenhall2020nerf} and its sparse-voxel variant~\cite{liu2020neural} were proposed to model the scene as a continuous 5D function that maps each spatial point to the radiance emitted in each direction and uses classical volume rendering techniques to render images. 
All these works focus only on static scenes, while our work targets at a dynamic setting with pose control, which is more challenging to model.

There are many recent works for dynamic scene rendering. 
\citet{Thies2019DeferredNR} assume that a coarse geometry of the scene is given and a neural texture is learned to synthesize novel views.
\citet{weng2020vid2actor} specifically studies free-view synthesis and pose control of a human from in-the-wild videos.
While the setup is very challenging, their results are still far from  video-realistic.
Neural Volumes~\cite{lombardi2019neural} and its follow-up work~\cite{wang2020learning} employ an encoder-decoder architecture to learn a compressed latent representation of a dynamic scene which synthesizes novel views by interpolation in the latent space and volume rendering.
Inspired by the recent success of neural radiance fields (NeRF), some works add a dedicated deformation network~\cite{tretschk2020nonrigid,park2020nerfies, pumarola2020dnerf}, scene flow fields~\cite{Li2020NeuralSF}, or space-time neural irradiance fields~\cite{xian2020spacetime} to handle non-rigid scene deformations.
\citet{peng2020neural} propose a set of latent codes attached to a body model in order to replay character motions from arbitrary view points.
\citet{lombardi2021mixture} introduce a mixture of volume primitives to avoid unnecessary sampling in empty space for dynamic scene rendering. 
\citet{su2021anerf} present an articulated NeRF representation based on a human skeleton for refining human pose estimation. 
Most of these works can only playback the same dynamic sequence of a scene under novel views. 
In contrast, we also model \textit{novel poses} under novel views, which is a much harder task because the network needs to generalize to new views \textit{and} to new poses.
~\citet{gafni2020dynamic} demonstrate the use of scene representations for synthesizing novel head poses and facial expressions from a fixed camera view and also generalization across identities has been demonstrated~\cite{raj2021pva}. 
Our work focuses on free-viewpoint synthesis of novel full human body poses. 
Compared to modeling the appearance and dynamics of a human face, modeling entire articulated humans for rendering and novel pose synthesis is a more challenging problem due to the articulated structure of the body, the appearance variations, self-occlusions, and highly  articulated motions. 
There are some concurrent  works~\cite{peng2021animatable,chen2021animatable}, which also propose a geometry-guided deformable NeRF for synthesizing humans in novel poses. However, these methods are not able to synthesize pose-dependent dynamic appearance.

%
\paragraph{Generative Models for Humans.}
Recently, generative adversarial networks (GANs) have made great progress in generating photorealistic images of humans and human body parts. 
Approaches that convert controllable conditioning inputs into photo-realistic body parts have been proposed for eyes \cite{Shrivastava2017}, hands \cite{Mueller2017}, and faces \cite{kim2018DeepVideo,tewari2020stylerig,tewari2020pie,GIF:3DV:2020}.
In the context of entire human bodies, many of the approaches formulate this task as an image-to-image mapping problem.
Specifically, these methods map the body pose in the form of rendering of a skeleton \cite{chan2019dance,SiaroSLS2017,Pumarola_2018_CVPR,KratzHPV2017,zhu2019progressive,kappel2020high-fidelity,Shysheya_2019_CVPR,li2019dense}, projection of a dense human model \cite{Liu2019,wang2018vid2vid,liu2020NeuralHumanRendering,Sarkar2020,Neverova2018,Grigorev2019CoordinateBasedTI,lwb2019, Raj_ANR, SMPLpix:WACV:2020}, or joint position heatmaps~\cite{MaSJSTV2017,Lischinski2018,Ma18} to realistic human images.
To better preserve the appearance from the reference image to the generated image, some methods~\cite{liu2020NeuralHumanRendering,Sarkar2020,yoon2021poseguided} first transfer the person's appearance from screen space to UV space and feed the rendering of the person in the target pose with the UV texture map into an image-to-image translation network.
Unlike these approaches, Textured Neural Avatar \cite{Shysheya_2019_CVPR} learns a person-specific texture map implicitly through backpropagation. 
To model stochasticity and the ability of sampling random human appearance, \cite{esser2018variational,Lassner2017,sarkar2021humangan} use the Variational Auto-Encoder \cite{Kingma2014VAE} framework conditioned on 2D pose.
All these methods do not learn scene geometry and cannot ensure multi-view consistency due to the 2D convolution kernels they used. 
Furthermore, the GAN-based neural rendering methods often show conspicuous ``shower curtain effects'', thus making them not suitable for free-viewpoint rendering. 
In contrast, our method learns multi-view consistent geometry that can be rendered by the standard ray marching method to perform consistent renderings across different camera views. 
\par
Recently, there are also works that explicitly or implicitly model the scene geometry.
\citet{wu2020multi} translate point clouds of the human performance into photoreal imagery from novel views. 
However, they can only replay the captured performance while we can also synthesize new performances.
\citet{habermann2021} jointly learn motion-dependent geometry as well as motion- and view-dependent dynamic textures from multi-view video input.
Although this method can produce high-quality results, it relies on a person-specific template which requires a 3D scanner and manual work for the rigging and skinning.
In contrast, our approach leverages a coarse parametric model -- SMPL which removes the need for a 3D scanner and manual work, and also supports reshaping of the actors body proportions for rendering.

\section{Neural Actor}
%
%
\begin{figure*}[t]
    \centering
    \includegraphics[width=\textwidth]{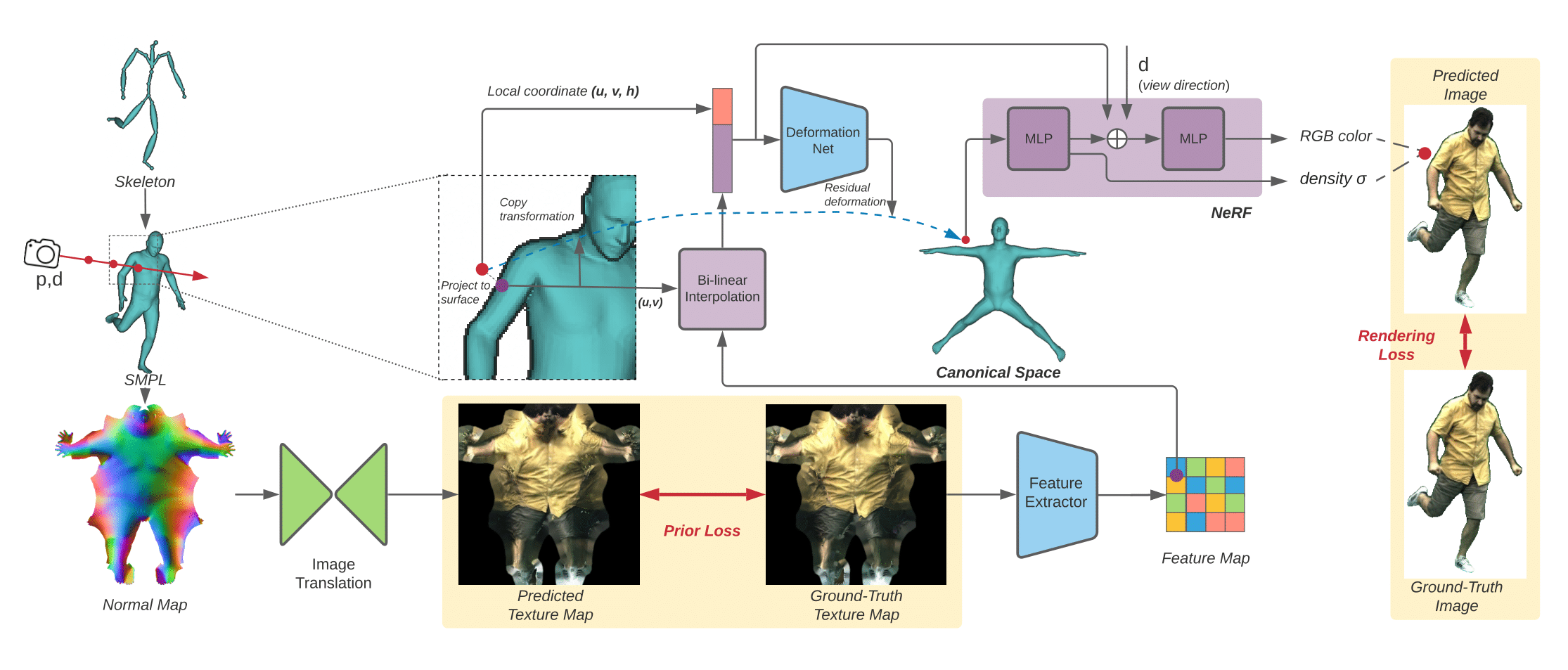}
    \caption{
    Overview of \emph{Neural Actor}. 
    Given a pose, we synthesize images by sampling points along camera rays near the posed SMPL mesh. 
    For each sampled point $x$, we assign to it the skinning weights of its nearest surface point and predict a residual deformation to transform $x$ to the canonical space. We then learn the radiance field in the canonical pose space to predict the color and density for $x$ using multi-view 2D supervision (\cref{sec.deformation}). 
    The pose-dependent residual deformation and color are predicted from the local coordinate of $x$ along with the texture features extracted from a 2D texture map of the nearest surface point of $x$. 
    At training time, we use the ground truth texture map generated from multi-view training images to extract the texture features. 
    At test time, the texture map is predicted from the normal map, which is extracted from the posed SMPL mesh via an image translation network, which is trained separately with the ground truth texture map as supervision (\cref{sec.texture}). 
    }
    \label{fig:pipeline}
\end{figure*}
%
%
\paragraph{Problem Formulation.}
Given a training set of $K$ synchronized RGB videos capturing a human actor with $T$ frames $\bm{\mathcal{I}}=\{\mathcal{I}_t^k\}$ along with its camera parameters $\bm{\mathcal{C}}=\{\mathcal{C}_t^k\}$ ($t=1\ldots T, k=1\ldots K$), our goal is to build an animatable virtual character with pose-dependent geometry and appearance that can be driven by arbitrary poses and rendered from novel viewpoints at test time. 
Note that we do not consider background synthesis in this paper and thus we apply color keying to extract the foreground in the images. 
Since body poses are needed as input, we track the body pose $\bm{\rho}$ for each frame. 

We first define a pose-conditioned implicit representation based on the state-of-the-art novel view synthesis method -- NeRF~\cite{mildenhall2020nerf} as follows:
%
\begin{equation}
    F_\theta: (\bm{x}, \bm{d}; \bm{\rho}) \rightarrow (\bm{c}, \sigma)
    \label{eq:nerf}
\end{equation}
%
%
where $\theta$ represents the network parameters. 
This function describes the color $\bm{c}=(r,g,b)$ and density $\sigma \in \mathbb{R}_+$ at spatial location $\bm{x}\in \mathbb{R}^3$ and view direction $\bm{d}\in \mathbb{S}^2$, conditioned on a pose vector $\bm{\rho}$.
Then, we apply the classical volume rendering techniques to render an image $\mathcal{I}$ of a human actor controlled by pose $\bm{\rho}$ with camera $\mathcal{C}$. 
Since this rendering process is differentiable, we can optimize $F_\theta$ by comparing $\mathcal{I}$ against the ground truth image without 3D supervision. 
%
%
\paragraph{Challenges.}
The first challenge is how to incorporate pose information into the neural implicit representation.
We observe that a na\"{i}ve design choice of $F_\theta$ by concatenating the pose vector $\bm{\rho}$ with $(\bm{x}, \bm{d})$ is inefficient for encoding a large number of poses into a single network for playback, and very difficult to generalize to novel poses (see \emph{NeRF+pose} in Figure~\ref{fig:qualicomp}). 
The second challenge is that learning dynamic geometric details and appearance of a moving human body from pose  only is an under-constrained problem, because at any moment the dynamic geometric details and changing appearance of a moving human body, such as cloth wrinkles, are not completely determined by the skeletal pose at that moment. 
Also, due to inevitable pose estimation errors, the association between dynamics and skeletal poses is even more difficult to learn. 
The above issues often lead to blurry artifacts in the output images, especially when using a deterministic model with maximum likelihood objectives, such as NeRF.
\par
To tackle the above challenges, NA improves the vanilla NeRF via \textit{template-guided neural radiance fields}. 
First, NA utilizes a deformable human body model (SMPL)~\cite{SMPL:2015} as a 3D proxy to deform implicit fields~(\cref{sec.deformation}). 
Second, to handle uncertainty in dynamic geometry and appearance, NA incorporates texture maps as latent variables~(\cref{sec.texture}) and thus can break down the mapping from body pose to dynamic effects into two parts, one from the pose to the texture map and the other from the texture map to dynamic effects. The former part can be trained with the adversarial loss to better mitigate the uncertainty issue. The texture map serves as a pose representation to predict pose-dependent effects in the latter part. 
An illustration of the overall pipeline is shown in Figure~\ref{fig:pipeline}.
%
%
\subsection{Geometry-guided Deformable NeRF}
\label{sec.deformation}
%
%
\paragraph{Deformation.}
Recent studies~\cite{park2020nerfies,pumarola2020d,tretschk2020nonrigid} have shown the effectiveness of representing dynamic scenes by learning a deformation function $\Phi_t(\bm{x}): 
\mathbb{R}^3\rightarrow\mathbb{R}^3$ to map every sample point $\bm{x}$ into a shared canonical space. 
By doing so, scenes across frames get connected through the canonical space as the common anchor, which improves training efficiency. 
However, restricted by the method design, it is difficult for these works to model relatively large movements efficiently and they show limited generalizability to novel poses.
%
%
To overcome these drawbacks, we augment this deformation function by querying an attached human body model -- SMPL~\cite{SMPL:2015}. 
SMPL is a skinned vertex-based model $(\mathcal{V},\mathcal{F},\mathcal{W})$ that represents a wide variety of body shapes in arbitrary human poses, where $\mathcal{V}\in\mathbb{R}^{N_V\times3}$ are the $N_V$ vertices, and $\mathcal{F}\in \{1\ldots N_V\}^{N_F\times3}$ are the vertex indices defining the triangles of the surface.
For each vertex $\bm{v}\in\mathcal{V}$, fixed skinning weights $\bm{\omega}\in \mathcal{W}$ are assigned, where $\sum_j\omega_j=1, \omega_j \geq 0, \forall j$.
Given a specific person (with fixed body shape), the SMPL model can be deformed according to the body pose vector $\bm{\rho}$ via Linear Blend Skinning~\cite{skinningcourse:2014}.
Since we want to transform the space in arbitrary poses to the canonical pose space, we perform an \emph{inverse-skinning transformation}~\cite{Huang:ARCH:2020} to deform the SMPL mesh in pose $\bm{\rho}$ to the canonical pose space:
%
%
\begin{equation}
    \Phi^{\text{SMPL}}(\bm{v}, \bm{\rho}, \bm{\omega}) = \sum_{j=1}^{N_J}\omega_j\cdot \left(R^j\bm{v}+\bm{t}^j\right),
    \label{eq.smpl}
\end{equation}
%
%
where $(R^j, \bm{t}^j)$ denotes the rotation and translation at each joint $j$ that transforms the joints back to the canonical space.
Equation~\eqref{eq.smpl} is only defined on the surface of SMPL, but can be extended to any spatial point in pose $\bm{\rho}$ by simply copying the transformation from the nearest point on the surface of SMPL. 
For any spatial point $\bm{x}$ in pose $\bm{\rho}$, we find the nearest point on the SMPL surface as follows:
%
%
\begin{equation}
    (u^*, v^*, f^*) = \arg\min_{u,v,f}\|\bm{x} - B_{u, v}\left(\mathcal{V}_{[\mathcal{F}(f)]}\right)
    \|_2^2,
    \label{eq.project}
\end{equation}
%
%
where $f \in \{1\ldots N_F\}$ is the triangle index, $\mathcal{V}_{[\mathcal{F}(f)]}$ is the three vertices of the triangle $F(f)$,  and $(u, v): u,v,u+v\in [0,1]$ represent the barycentric coordinates on the face.
$B_{u,v}(.)$ is the barycentric interpolation function. 
%
%
%
\par 
Next, we model the pose-dependent non-rigid deformation which cannot be captured by standard skinning using a residual function $\Delta\Phi_\theta(\bm{x}, \bm{\rho})$, similar to \cite{pumarola2020d,tretschk2020nonrigid}.
The full deformation model can be represented as:
%
%
%
\begin{equation}
    \Phi_\theta(\bm{x},\bm{\rho}) = \Phi^{\text{SMPL}}(\bm{x}, \bm{\rho}, \bm{\omega}^*) + \Delta\Phi_\theta(\bm{x}, \bm{\rho}),
    \label{eq.deform}
\end{equation}
%
%
where $\bm{\omega}^*=B_{u^*, v^*}\left(\mathcal{W}_{[\mathcal{F}(f^*)]}\right)$ are the corresponding skinning weights of the nearest surface point.
The full model allows us to pose the mesh via skinning as well as to model non-rigid deformations with the residual function.
With this design, learning dynamic geometry becomes more efficient since we just need to learn a residual deformation for each pose. 
Also, $\Delta\Phi_\theta(\bm{x}, \bm{\rho})$ serves to compensate  unavoidable tracking errors in  marker-less 
motion capture.
%
%
\paragraph{Rendering.}
Once points are deformed into the canonical space, we learn NeRF in this space following Equation~\eqref{eq:nerf}. 
The final pixel color is predicted through volume rendering~\cite{kajiya1984ray} with $N$ consecutive samples $\{\bm{x}_1,\ldots\bm{x}_N\}$ along the ray $\bm{r}$:
%
%
\begin{equation}
    \mathcal{I}(\bm{r}, \bm{\rho})= \sum_{n=1}^N
    \left(\prod_{m=1}^{n-1}
    e^{-\sigma_m \cdot \delta_m}
    \right)\cdot\left(1-e^{-\sigma_n \cdot \delta_n}\right)\cdot\bm{c}_n,
    \label{eq.rendering}
\end{equation}
%
%
where $\sigma_n=\sigma(\Phi_\theta(\bm{x}_n,\bm{\rho}))$, $\bm{c}_n=\bm{c}\left(\Phi_\theta(\bm{x}_n,\bm{\rho}), \bm{d}, \bm{\rho}\right)$ and $\delta_n=\|\bm{x}_n-\bm{x}_{n-1}\|_2$.
Note that we use only the deformed points to estimate densities ($\sigma$) to enforce learning the shared space, while including pose $\bm{\rho}$ to predict colors ($\bm{c}$) with pose-dependent phenomena (e.g. shadows). Here, we do not use the pose vector as input to NeRF directly. Instead, we use the texture map as input serving as a local pose representation to better infer dynamic effects, which we will elaborate later. 
%
%
\par
\begin{figure}[t]
    \centering
    \includegraphics[width=0.6\linewidth]{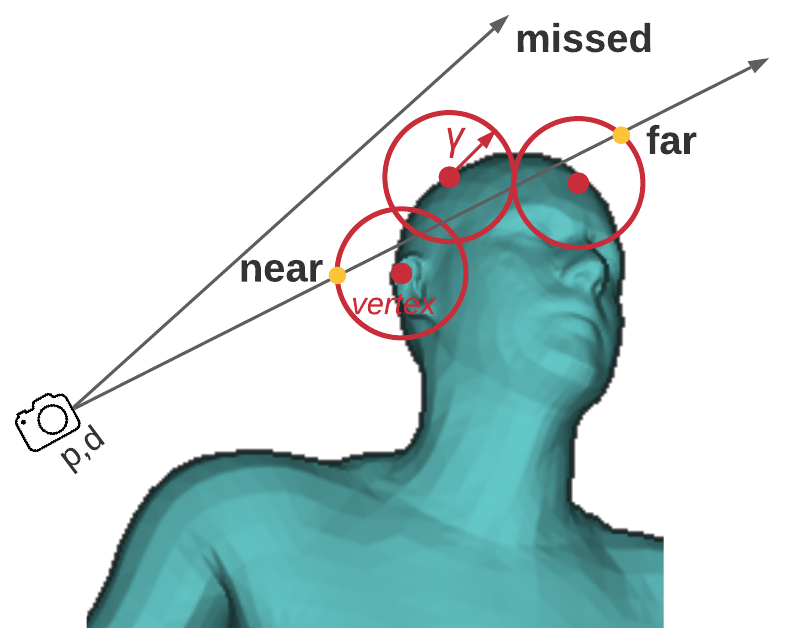}
    \caption{Illustration of geometry-guided ray marching.}
    \label{fig:example_intersect}
\end{figure}

 
The vanilla NeRF uses a hierarchical sampling strategy: 
the second stage samples more points where the initial uniform samples have a higher probability. 
We interpret it as sampling based on the geometry learned in the first stage. 
In our setting, since the SMPL mesh is given, we adopt a geometry-guided ray marching process to speed up the volume rendering process.
As shown in Figure~\ref{fig:example_intersect}, we take uniform samples but only accept samples $\bm{x}$ if $\min_{\bm{v}\in\mathcal{V}}\|\bm{x} - \bm{v}\|_2 < \gamma$, where $\gamma$ is a hyperparameter which defines how close SMPL approximates the actual surface. More implementation details can be found in Appendix. 
%
%

\subsection{Texture Map as Latent Variables}
\label{sec.texture}
NeRF is only capable of learning a deterministic regression function, which makes it not suitable for handling uncertainty involved in modeling dynamic details.
As mentioned earlier, the mapping from the skeletal pose to dynamic geometry and appearance is not a bijection. 
Consequently, direct regression often leads to blurry outputs (see `NA w/o texture' in Figure~\ref{ablation1}). 
A common approach is to incorporate latent variable $\bm{z}$, i.e. $p(\sigma, \bm{c}|\bm{\rho})=\int_{\bm{z}}p(\sigma, \bm{c}|\bm{z},\bm{\rho})\cdot p(\bm{z}|\bm{\rho})d\bm{z}$. 
For example, we can choose $\bm{z}$ as spherical Gaussian and model the NeRF output $(\sigma, \bm{c})$ using conditional VAEs~\cite{Kingma2014VAE,lombardi2019neural}.
%
\par
In contrast to common choices, we take the full advantage of the SMPL template and learn structure-aware latent variables.
Specifically, we take a 2D texture map $\mathcal{Z}\in \mathbb{R}^{H\times W\times C}$ as the latent variable, which is defined based on a fixed UV parameterization $\mathcal{A}\in [0,1]^{N_F\times3\times2}$ which maps points on the 3D mesh surface to a 2D UV plane.
There are three advantages of choosing such $\mathcal{Z}$:
%
%
\begin{enumerate}[leftmargin=*]
    \item{
        Compared to a compressed representation (e.g. latent vectors used in \cite{lombardi2019neural}), the texture map has higher resolution, making it possible to capture local details. 
        The local information can be used as a local pose representation to infer the local geometry and appearance changes in the scene representation. Furthermore, this local pose representation, compared to a global pose representation (e.g. a pose vector), facilitates the generalizability of NA to novel poses.
    }
    %
    %
    \item{
        A simple posterior $q(\mathcal{Z}|\bm{\mathcal{I}},\bm{\rho})$ is available. 
        That is, during training, the texture map $\mathcal{Z}$ for each training frame can be obtained by back-projecting the training images of each frame to all visible vertices and generate the final texture map by calculating the median of the most orthogonal texels from all views, as done  in~\citet{alldieck2018video}.  
        As we do not need to update $q$, learning of $p(\sigma, \bm{c}|\bm{\rho})$ can be readily split into two parts, learning of the prior $p(\mathcal{Z}|\bm{\rho})$, and learning of the rendering $p(\bm{\mathcal{I}}|\mathcal{Z},\bm{\rho})$. 
    }
    %
    %
    \item{
        Inspired by~\citet{liu2020NeuralHumanRendering}, the learning of the prior model $p(\mathcal{Z}|\bm{\rho})$ can be formulated as an image-to-image translation task which maps normal maps generated from the posed meshes to texture maps. 
        Fortunately, such a problem has been well-studied in the literature. To better preserve temporal consistency, we use vid2vid~\cite{wang2018vid2vid} to predict high-resolution texture maps from normal maps. The adversarial loss in vid2vid prevents the training from converging to the mean value of $\bm{\mathcal{I}}$  and thus mitigates the uncertainty issue. 
    }
\end{enumerate}
%
%
As shown in Figure~\ref{fig:pipeline}, we apply an additional feature extractor $G(.)$ after $\mathcal{Z}$ to extract high-level features of the surface appearance which contain significantly more information than the RGB values of the texture maps.
For any spatial point $x$, its pose-dependent local properties depend on the extracted features of $\mathcal{Z}$ at its nearest surface point searched through Equation~\eqref{eq.project} and its local coordinate $(u,v,h)$ where $(u, v)$ is the texture coordinate of the nearest surface point and $h$ is the signed distance to the surface. 
The feature extractor is trained together with the geometry-guided fields (\cref{sec.deformation}) for predicting both residual deformations and dynamic appearance.
%

%
%

%
%

\section{Experiments}
%
%
%

\paragraph{Datasets.}

To validate our approach, we tested on eight sequences from three different datasets, including one captured by ourselves, which contain different actors wearing various textured clothing. 
We used two sequences, $S1$ and $S2$, from the \emph{DeepCap} dataset~\cite{deepcap} with $11$ and $12$ cameras, respectively, at a resolution of $1024 \times 1024$. $S1$ contains $38{,}194$ training frames and $23{,}062$ testing frames;  $S2$ has $33,605$ training frames and $23{,}062$ testing frames. We also evaluated on two sequences, $D1$ and $D2$, from the \emph{DynaCap} dataset~\cite{habermann2021}, under a dense camera setup at a resolution of $1285 \times 940$. The two sequences have approximately $20{,}000$ and $7{,}000$ frames for training and testing, respectively. For the \emph{D1} sequence, we used $43$ cameras for training and $4$ uniformly distributed held-out cameras for the evaluation of our method on novel camera views; for the \emph{D2} sequence, we used $100$ cameras for training. 

To further evaluate our method on a wider variety of body poses and more challenging textured clothing, we captured a new multi-view human performance corpus with $79-86$ cameras at a resolution of $1285 \times 940$. It contains four sequences, $N1$-$N4$, and each has $12{,}000-16{,}000$ frames for training and around $8,000$ frames for testing. All the actors have given consent in signed written forms to the use of their recordings and synthesized results in this work. We will make the dataset publicly available. See Figure~\ref{fig:dataset} in the Appendix for the detailed information of the training data. 

In addition, to demonstrate the generalizability of the proposed method, we additionally tested our method with various dancing motions from the \emph{AIST} dataset~\cite{aist-dance-db,li2021learn} and very challenging motions from the \emph{AMASS} dataset~\cite{AMASS:ICCV:2019} as the driving poses. Note that these poses are quite distinct from the training poses, thus making the reenactment task challenging.
%

\paragraph{Data Processing.} 
Since we are only interested in foreground synthesis, we use color keying to extract the foreground in each image. 
We then employ an off-the-shelf SMPL tracking system\footnote{\url{https://github.com/zju3dv/EasyMocap}} to optimize the shape parameters of SMPL as well as the global translation and the SMPL's pose parameters (a 72-dimensional  vector). The pose parameters include the root orientation and the axis-angle representations of the relative rotations of 23 body parts with respect to its parent in the kinematic tree. We further normalize the global translation for camera positions and the tracked geometry in our model as well as the baseline models. 

We follow the standard texture generation step in~\cite{alldieck2018video} to generate ground truth texture maps for training the image translation network.
In our early development, we observed in the experiments that the boundary pixels in the UV space cannot preserve continuity in the 3D space, which could affect the texture feature extraction stage in our method.  We take two measures to alleviate this problem. First, we cut the seam of the SMPL mesh in Blender and unwarp the mesh into one piece in the UV space. Second, we perform inpainting on the dilated region of the generated texture maps. An example of the resulted texture map is presented in Figure~\ref{fig:pipeline}.



\begin{figure*}[h]
    \centering
    \includegraphics[width=\linewidth]{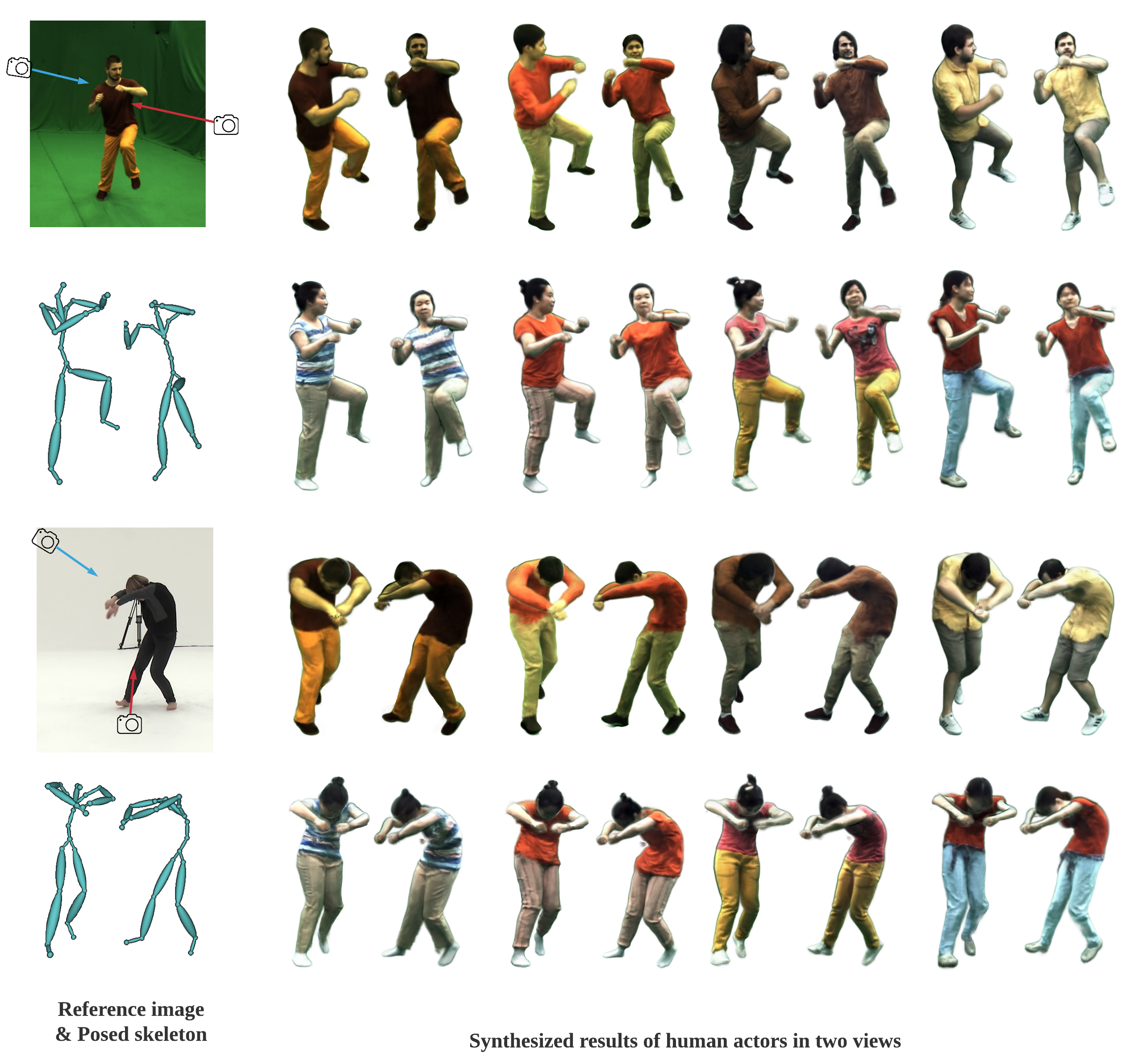}
    \caption{\label{Qualitative1}
    Qualitative reenactment results with the driving poses from the \emph{DeepCap} testing set~\cite{deepcap} and the \emph{AIST} dataset~\cite{aist-dance-db, li2021learn}. 
    Note that our method can synthesize photorealistic images of human characters even for the unseen poses and views that strongly differ from the training poses.
    Courtesy of \citet{aist-dance-db} for the AIST reference image.
    }
\end{figure*}

\paragraph{Implementation Details.} 
We model the residual deformation networks $\Delta\Phi$ as 2-layer MLPs, and follow the network design of  NeRF\\~\cite{mildenhall2020nerf} to predict density and color of each spatial location in the canonical space. 
We apply positional encoding to spatial location with a maximum frequency of $L=6$. 
The texture map $\mathcal{Z}$ is at a resolution of $512\times 512$. 
We use the backbone of ResNet34, which was pre-trained on ImageNet, as the texture feature extractor $G(.)$ to extract features from texture maps. 
We extract feature maps prior to the first 4 pooling layers, upsample them using bilinear interpolation and concatenate them to form multi-level features as the output $G(\mathcal{Z})$ in $256\times256\times512$, similar to~\cite{yu2020pixelnerf}.
A detailed illustration of the proposed architecture is shown in Figure~\ref{fig:architecture}.

The texture feature extractor is trained together with the residual deformation network as well as NeRF in the canonical space with the L2 loss measuring the difference between the rendered images and ground truth images. 
The training takes around 2 days on $8$ Nvidia V100 32G GPUs for $300$K iterations with a batch size of $1024$ rays per GPU. 
For learning the prior, we used  vid2vid~\cite{wang2018vid2vid}\footnote{\url{https://github.com/NVlabs/imaginaire}} with the default setting to predict texture maps at $512 \times 512$ pixels from normal maps in $512 \times 512$. We trained vid2vid on $4$ Nvidia Quadro RTX 8000 48G GPUs with batchsize $4$ per GPU for about $10$K iterations for around 3 days. 
Since these two steps are independent, we can train them in parallel. 
At test time, rendering a 940x1285 image takes around 4 seconds, and 6-8G GPU memory. 

%

\begin{figure*}[h]
    \centering
    \includegraphics[width=0.9\linewidth]{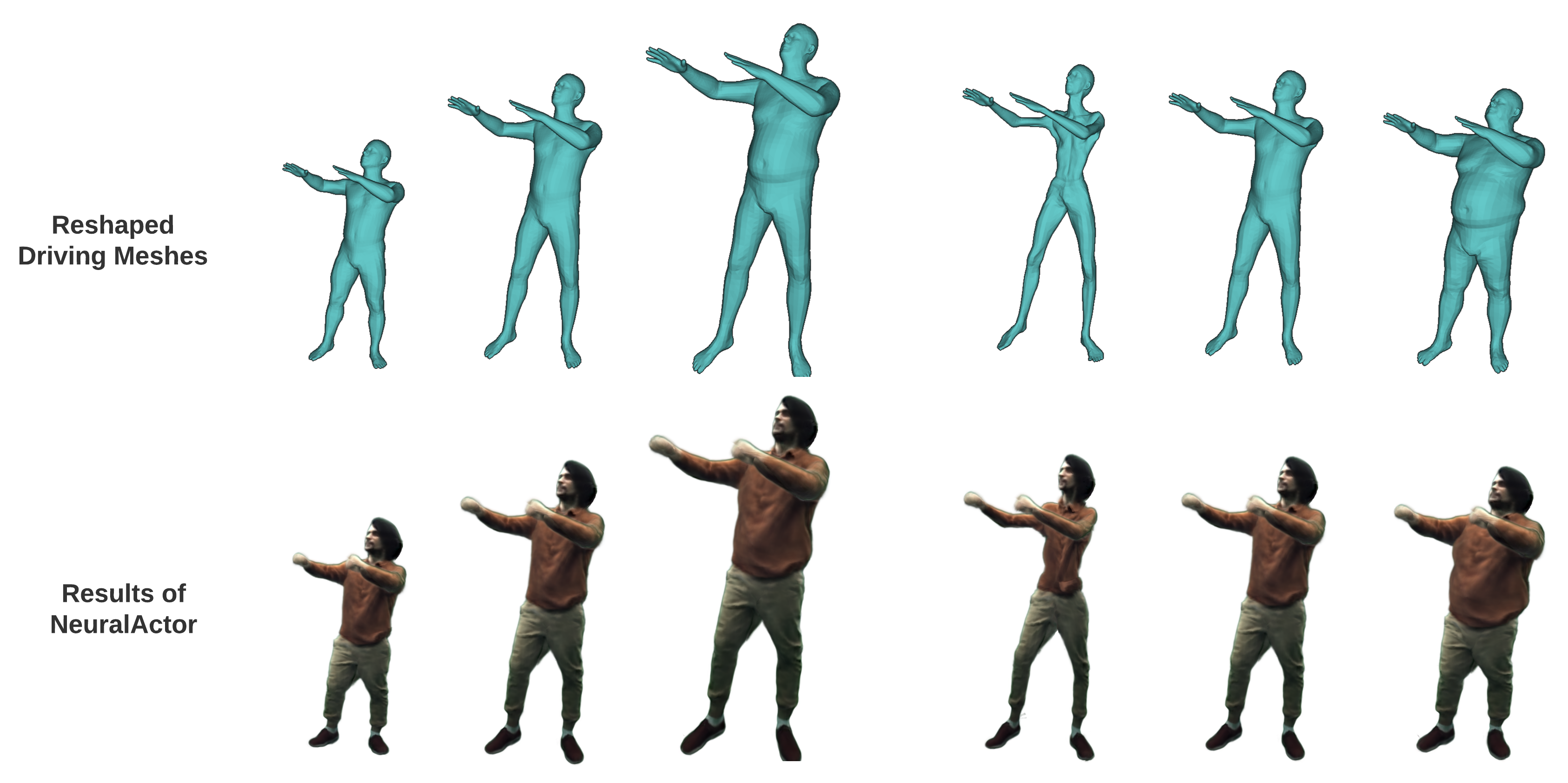}
    \caption{\label{Reshape} Rendering results of our method for different body shape configurations of the same actor. 
    Note that our method can also produce photorealistic results for shapes that strongly differ from the original shape of the actor.
    }
    \label{fig:reshape}
\end{figure*}

\begin{figure}[h]
\centering
\includegraphics[width=\linewidth]{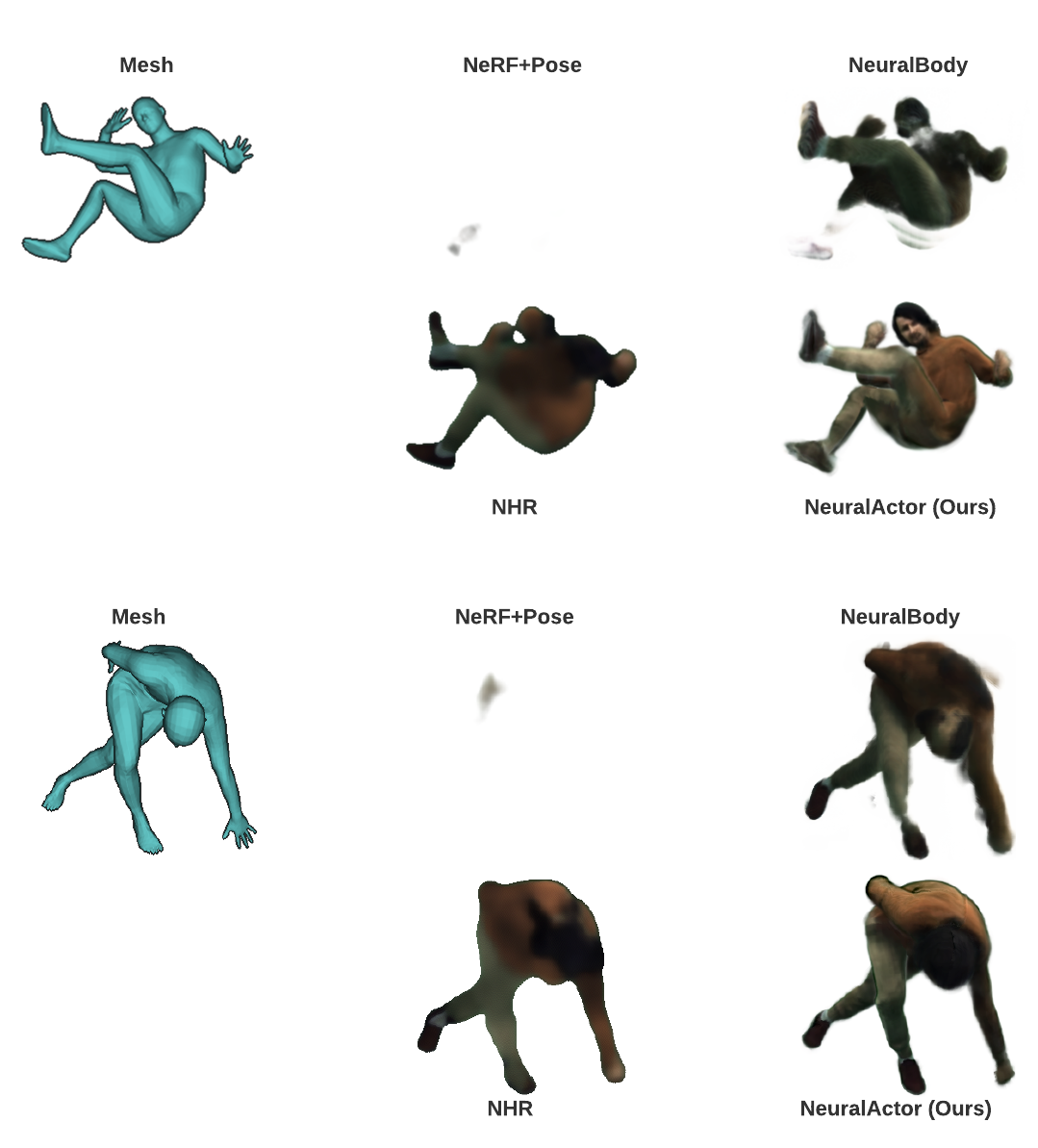}
\caption{\label{Challenge}Comparison on novel pose synthesis with very challenging poses.}
\end{figure}

\subsection{Qualitative Results}
Since our model only requires the posed SMPL mesh as condition for novel view synthesis, it can easily perform applications such as \textit{reenactment} and \textit{body reshape}.
\paragraph{Reenactment.} 
We directly use the pose parameters from the driving person and the shape parameters from the target person to get the posed SMPL mesh. Figure~\ref{Qualitative1} shows example reenactment results where we use the testing poses from the \emph{DeepCap} dataset~\cite{deepcap} and the AIST dataset~\cite{aist-dance-db,li2021learn} as driving poses, respectively. 
We can see that our method can synthesize faithful imagery of humans with fine-scale details in various motions and generalize well to challenging motions. 
In addition, we further test our approach on some very challenging poses, such as crunch, bending forward. As shown in Figure~\ref{Challenge}, our approach is able to produce plausible synthesized results of such challenging poses, which significantly outperforms the baseline methods. Note that NV cannot perform reenactment where a person drives a different person since it requires the captured images as input at test time.

\paragraph{Body Reshape.} 
As shown in Figure~\ref{fig:reshape}, we can adjust the shape parameters (PC1 and PC2) of the SMPL template  to synthesize animations of the human in different shapes. 
Specifically, at inference time, we first warp the posed space to the canonical pose space for the reshaped human template via inverse kinematic transformation, and then transform the canonical pose space of the reshaped template to that of the original shape template and finally infer the color and density in the original canonical space. 
This technique will be potentially useful for the movie industry, e.g. we are able to synthesize animations of a giant or dwarf by modifying the shape parameters of any actor, without the need of finding the actual human in that shape. Please refer to the supplemental videos for more results.

%
%

%
%
\subsection{Comparisons}
%
%
\begin{figure*}[t]
    \centering
    \includegraphics[width=\linewidth]{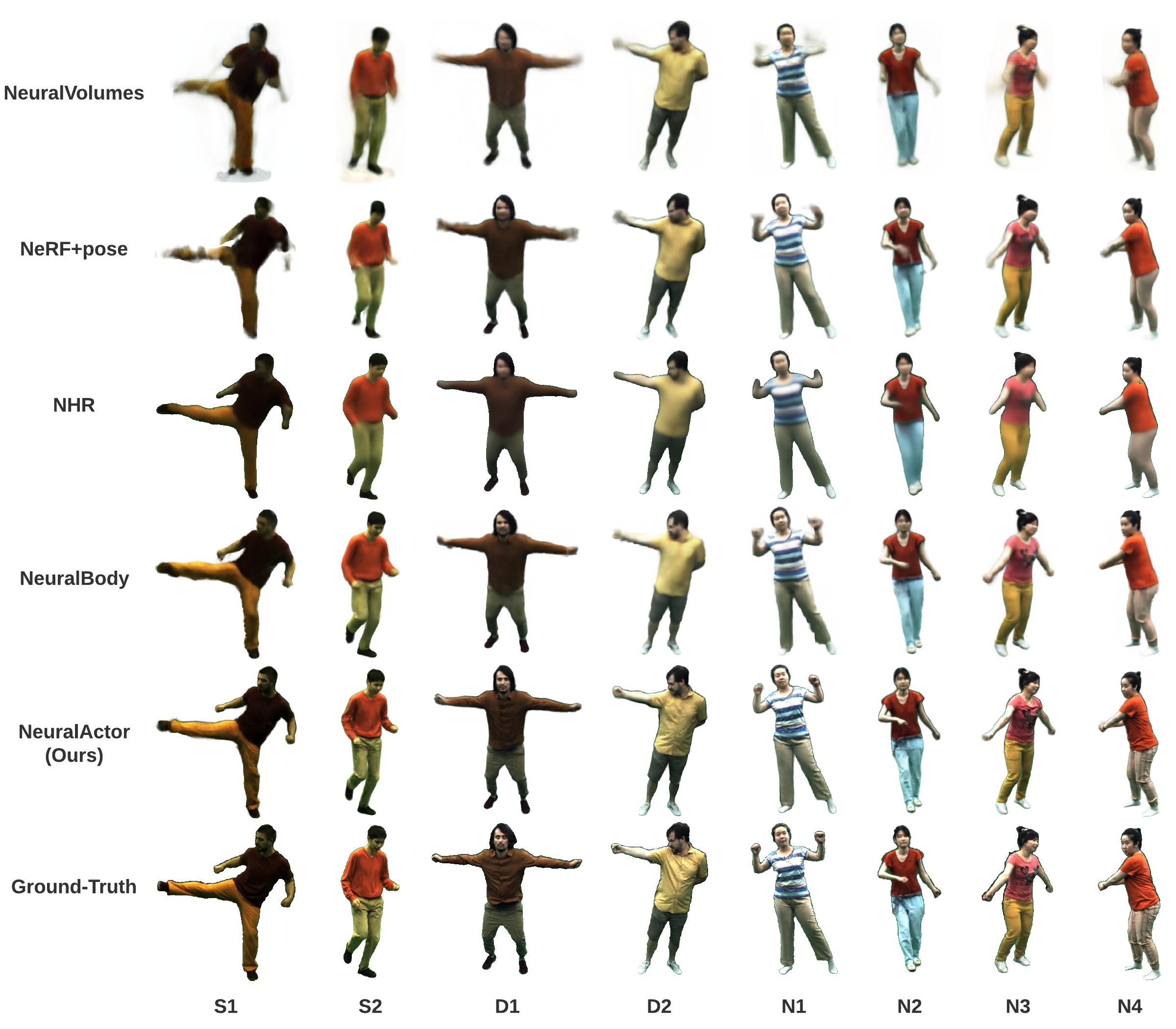}
    \caption{\label{Qualitative_novel_pose_compare}
    Qualitative comparisons on novel pose synthesis with eight sequences. 
    Our method can faithfully recover the pose-dependent wrinkles and appearance details which cannot be achieved by other baseline methods. 
    }
    \label{fig:qualicomp}
\end{figure*}
We validate the proposed method on two tasks: novel view synthesis and novel pose synthesis, comparing with recent baselines.

\paragraph{Novel Camera View Synthesis.} For this comparison, we evaluate on the \emph{D1} sequence, where 43 cameras are used for training and 4 uniformly distributed held-out cameras are used for evaluation. 
We compare our method with four state-of-the-art neural rendering methods.
More precisely, we compare to: 
\begin{itemize}[leftmargin=*]
    \item \emph{NeRF+pose}: We extend NeRF~\cite{mildenhall2020nerf}, a state-of-the-art view synthesis method for static scenes, to a pose-conditioned NeRF by feeding a pose vector into the vanilla NeRF directly.
    \item \emph{Neural Volumes (NV)~\cite{lombardi2019neural}}: NV utilizes an encoder-decoder architecture to learn a latent representation of a dynamic scene that enables it to sample or interpolate a latent vector to produce novel content. 
%
%
We follow the original setting of NV, and provide the images captured from three uniformly distributed cameras at both training and test stages for each pose to encode the content into the latent space. 
\item \emph{Neural Body (NB)~\cite{peng2020neural}}: NB extends the vanilla NeRF by utilizing sparseCNNs to encode spatial features from the posed mesh. We follow the original setting of NB. 
\item \emph{Multi-View Neural Human Rendering (NHR)~\cite{wu2020multi}}: NHR extracts 3D features directly from the input point cloud and project them into 2D features. We use the vertices of the SMPL model as the input point cloud. 
\end{itemize}
\begin{table}[h]
   \small
   \centering
      \caption{\label{novel_view_compare} Novel camera view synthesis on the training poses of D1}
     \begin{tabular}{lcccc}
      \toprule
            Models & PSNR$\uparrow$ & SSIM$\uparrow$ & LPIPS$\downarrow$ & FID$\downarrow$\\
      \midrule
      NeRF + pose & 22.791  & 0.921 & 0.156 & 146.135 \\
      NV & 24.248	& 0.924	 &0.149	& 131.86\\
      NB & 24.447&0.934	& 0.116	& 119.04\\
      NHR & 22.587 & 0.928 & \bf{0.072} & 164.85\\
      NA (Ours) & {\bf 24.875} & {\bf 0.941}& 0.079 & \bf{45.649}\\
      NA w. GT (Ours) & {\bf 27.567} & {\bf 0.959} & \bf{0.071} & \bf{43.089} \\
      \midrule
     \end{tabular}

\end{table}
\begin{table*}[t]
\caption{\label{Quantitative_novel_pose_compare}The quantitative comparisons on test poses of eight sequences. We use three metrics: PSNR, FID~\cite{heusel2017gans} and LPIPS~\cite{zhang2018perceptual} to evaluate the rendering quality. 
To reduce the influence of white background,  all the scores are calculated from the images cropped with a maximum 2D bounding box which is estimated from the foreground masks of all the target images.
The scores are averaged over all the training views of every $10^\text{th}$ test poses. Note that, since PSNR is a metric based on least squares measurement, it does not faithfully measure image sharpness and so cannot properly account for the nuances of human visual perception.~\cite{zhang2018perceptual}.  } %
\begin{center}
\small
\scalebox{1.0}{
\begin{tabular}{lcccccccccccc}
\toprule
  & \multicolumn{3}{c}{\emph{D1}}  &  \multicolumn{3}{c}{\emph{D2}} & \multicolumn{3}{c}{\emph{S1}} & \multicolumn{3}{c}{\emph{S2}} \\ 
  \midrule
  Models & PSNR$\uparrow$ & LPIPS$\downarrow$ &FID $\downarrow$ & PSNR$\uparrow$  & LPIPS$\downarrow$ &FID $\downarrow$ &PSNR$\uparrow$ & LPIPS$\downarrow$ & FID$\downarrow$ &PSNR$\uparrow$ & LPIPS$\downarrow$ & FID$\downarrow$\\
\midrule
 NeRF + pose  &
 22.791	& 0.156 & 146.14 &
 23.339	& 0.123	& 134.22 &
 22.328	& 0.158	& 126.44 &
 23.445	& 0.134	& 112.11 \\
 NV &
 20.648 & 0.171	& 135.57 &
 21.020	& 0.143	& 122.36 & 
 18.661	& 0.190	& 123.04 &
 19.076	& 0.173	& 98.063 \\
 NB &
 \bf{23.768}	& 0.119	& 117.73 &
 \bf{23.872}	& 0.112	& 124.39 &
 \bf{22.967}	& 0.114	& 92.098 &
 \bf{23.946}	& 0.096	& 81.527 \\ 
 NHR &
 22.237	& \bf{0.075}	& 162.62 &								
 22.997	& 0.070	& 138.25 &
 14.530	& 0.217	& 124.56 &
 22.419	& 0.073	& 149.16 \\ 
NA (Ours) &
23.547 & 0.084 & \bf{44.921} & 
23.785 &  \bf{0.065} & \bf{46.812} &
22.495 &  \bf{0.084} & \bf{34.361} &
23.531 &  \bf{0.066} & \bf{19.714} \\
 \midrule
 \midrule
   & \multicolumn{3}{c}{\emph{N1}}  &  \multicolumn{3}{c}{\emph{N2}} & \multicolumn{3}{c}{\emph{N3}} & \multicolumn{3}{c}{\emph{N4}} \\
  \midrule
  Models & PSNR$\uparrow$ & LPIPS$\downarrow$ &FID $\downarrow$ & PSNR$\uparrow$  & LPIPS$\downarrow$ &FID $\downarrow$ &PSNR$\uparrow$ & LPIPS$\downarrow$ & FID$\downarrow$ &PSNR$\uparrow$ & LPIPS$\downarrow$ & FID$\downarrow$\\
 \midrule
 NeRF + pose  &
 22.892	& 0.174	& 125.83 &
 23.922	& 0.142	& 126.34 &
 24.621	& 0.113	& 106.30 &
 23.648	& 0.162	& 153.30 \\
 NV &
 20.901	& 0.183	& 104.13 &  
 21.372	& 0.155	& 100.82 &
 21.394	& 0.132	& 108.42 &
 20.617	& 0.181	& 115.00 \\
 NB &
 \bf{23.159}	& 0.153	& 107.72 &
 \bf{24.006}	& 0.115	& 91.218 &
 \bf{25.273}	& 0.093	& 76.124 &
 \bf{24.192}	& 0.140	& 111.78 \\ 
 NHR &
21.630 & 0.098	& 117.42 &
22.806 &  \bf{0.075}	& 175.00 &
23.719 & 0.062	& 91.535 &
22.744 & 0.092	& 164.23 \\ 
NA (Ours) &
22.799 &  \bf{0.084} & \bf{30.218} & 
24.345 & 0.080 & \bf{39.918} & 
25.014 &  \bf{0.057} & \bf{25.946}	& 
23.861 &  \bf{0.079} & \bf{28.525} \\
\bottomrule
\end{tabular}}
\end{center}
\end{table*}

We show the quantitative results in  Table~\ref{novel_view_compare} and include the visual results in the supplemental video. 
For all the baseline methods, it is difficult to perform photo-realistic rendering for playback when the training set contains a large number of different poses, e.g. pose sequences with ~20K frames. 
Note that \emph{NV} and \emph{NB} have demonstrated good results in their work for playing back a short sequence, e.g. 300 frames, however, encoding a large number of frames, e.g. ~20K frames into a single scene representation network tends to produce blurriness in the results due to the large variations in the training data. 
Simply feeding pose vectors into NeRF (NeRF+pose), which was similarly used in~\cite{Gafni20arxiv_DNRF}), is not efficient for training since full deformations need to be learned. 
Furthermore, NeRF+pose produces blurry artifacts due to the uncertainty in the mapping from the skeletal pose to dynamic geometry and appearance.
\emph{NHR} also has difficulties in encoding a large number of poses and leads to blurry results. 
In contrast, our method improves the training efficiency and resolves the blurriness issue by disentangling the full deformation into inverse kinematic transformation and residual deformation and learning a prior with texture maps to resolve the blurriness issue. 
With these strategies, we can synthesize high-quality renderings of humans with sharp dynamic appearance for the playback of a long sequence. 
We note that our results can further be improved when the multi-view images are provided for generating the `Ground Truth' texture maps at test time (see \emph{NA w. GT}). 
%

%
%
%
%
%

%

\paragraph{Novel Pose Synthesis.} For novel pose synthesis, we first conduct a comparison with the above four baselines on eight sequences, where the testing poses are used for evaluation. 
The qualitative and quantitative results are reported in Figure~\ref{Qualitative_novel_pose_compare} and Table~\ref{Quantitative_novel_pose_compare}, respectively. 
\emph{NeRF+pose} and \emph{NV} produce severe artifacts in the results, such as missing body parts and blurriness. 
\emph{NB} and \emph{NHR} also suffer from blurriness and cannot preserve dynamic details in the results.
Our method can generalize to new poses well and achieve high-quality results with sharp details which are significantly better than the baseline methods.  
Our method proposes three main design choices that lead to an improvement over the baselines: 1) NA disentangles body movements into inverse skinning transformations and dynamic residual deformations where only the latter needs to be learned, thus facilitating the training efficiency; 2) By incorporating 2D texture maps as latent variables, NA effectively mitigates the uncertainty in the mapping from the skeletal pose to dynamic geometry and appearance; 3) The local features extracted from the high-resolution texture maps serve as a local pose representation for inferring local changes in geometry and appearance. This local representation not only enables better capturing of geometric details but also makes the model generalize well to new poses.

\begin{figure}[h]
    \centering
    \includegraphics[width=0.9\linewidth]{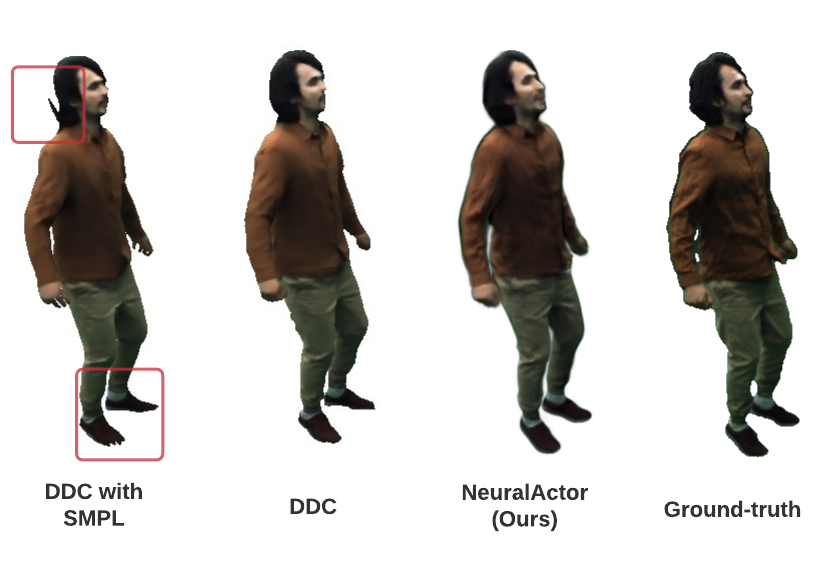}
    \caption{Comparison to DDC.
    While the template-based approach DDC produces  high-quality results, it requires a personalized 3D scan of the actor and manual work is needed for the rigging and skinning.
    To bring their setting closer to ours, we also compare to DDC where we replace the template with the SMPL model.
    Note that simply applying their method with a SMPL model results in geometric artifacts.
    In contrast, our method achieves a similar quality to their original method without requiring a personalized template.
    }
    \label{fig:ddc}
\end{figure}
We further compare with a recent mesh-based method, Real-time Deep Dynamic Characters (\emph{DDC})~\cite{habermann2021}, on the \emph{D1} sequence. The original \emph{DDC} requires a person-specific template captured by a 3D scanner. 
Since our method only needs the SMPL model, as requested, a comparison has been conducted with the SMPL model as input (\emph{DDC} with \emph{SMPL}). 
We also provided the original result of \emph{DDC} with a  person-specific template for reference. 
As shown in the Figure~\ref{fig:ddc}, \emph{DDC} works well with a person-specific template, however, deforming a coarse SMPL mesh is more challenging, which leads to artifacts on the deformed geometry, such as the head. 

Our method is also related to Textured Neural Avatar (\emph{TNA}).
However, because its code and data are not available, we will just conceptually discuss the difference with that work.
Different from our method, \emph{TNA} is unable to synthesize dynamic appearance of humans. Moreover, their results are not view-consistent and often suffer from artifacts such as missing body parts (see supplemental video).
%
%
%
%
\begin{figure*}[t]
   \begin{minipage}{0.47\textwidth}
   \small
     \begin{tabular}{lcccc}
      \toprule
      Models & PSNR$\uparrow$ & SSIM$\uparrow$& LPIPS$\downarrow$ & FID$\downarrow$\\
      \midrule
      NA (Full model) & 23.547 & 0.928&{\bf 0.084} &{\bf 44.921}\\
      \; {w/o texture inputs} & {\bf 24.181} & \bf{0.930} &0.131 &108.30 \\
      \; {raw texture inputs} & 23.110 &0.916& 0.142  & 105.45 \\
      \; {normal map inputs} & 19.316 &0.887& 0.167  & 148.56 \\
      \bottomrule
     \end{tabular}
   \end{minipage}
   \begin{minipage}{0.51\textwidth}
     \includegraphics[width=\linewidth]{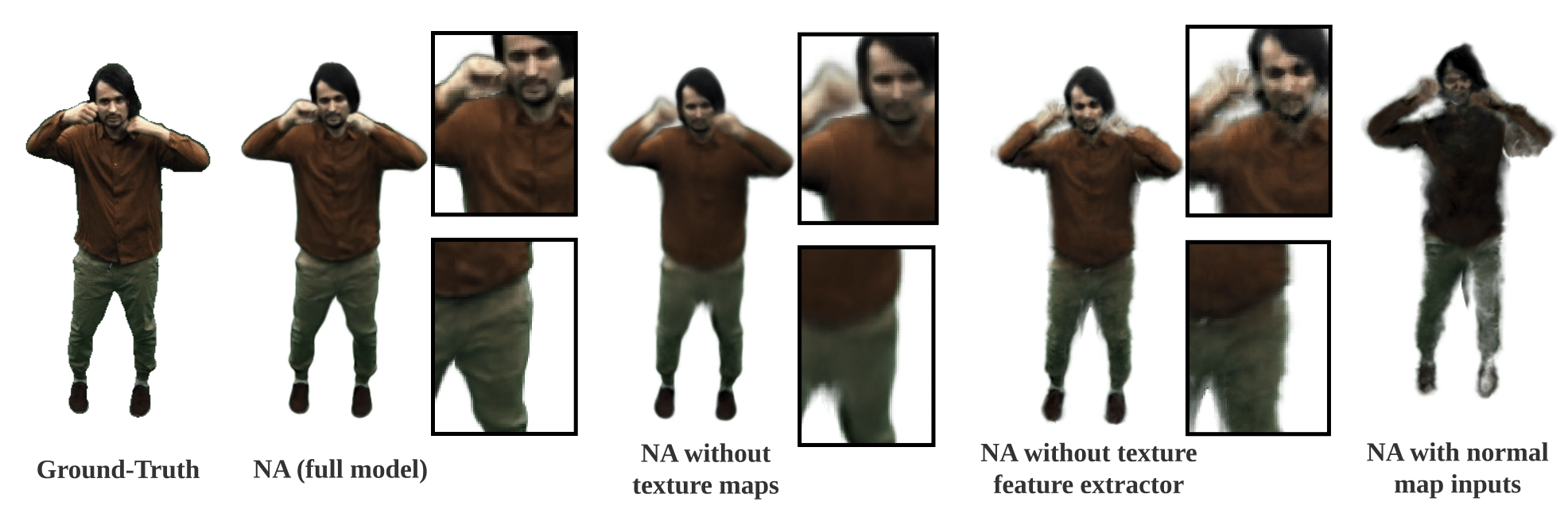} 
   \end{minipage}
   \caption{\label{ablation1} Ablation on using texture features as latent variables. (Left) Quantitative results; (Right) Visual comparison.}
\end{figure*}
\begin{figure*}[h]
   \begin{minipage}{0.47\textwidth}
   \small
     \begin{tabular}{lcccc}
      \toprule
      Models & PSNR$\uparrow$ & SSIM$\uparrow$ & LPIPS$\downarrow$ & FID$\downarrow$\\
      \midrule
      NA (Full model) & 23.547 & 0.928 &{\bf 0.084} &{\bf 44.921} \\
      \; {w/o residual deformation} & 23.532 &0.926&	0.093 & 56.580\\
      \; {w/o geometry guidance} & 21.635 &	0.909 & 0.137 &	72.379 \\
      \midrule
      \; {using nearest vertex} & \bf{23.625} & \bf{0.930} & 0.092   & 70.768 \\
      \bottomrule
     \end{tabular}
   \end{minipage}
   \begin{minipage}{0.51\textwidth}
     \includegraphics[width=\linewidth]{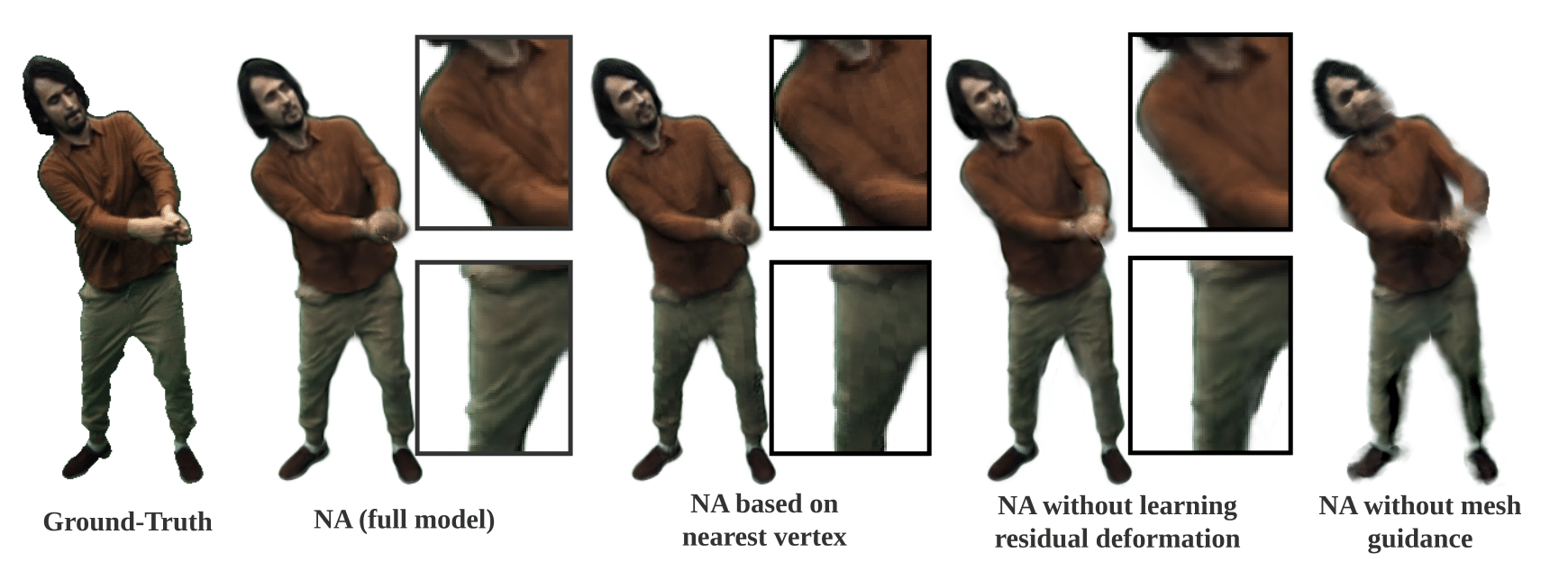} 
   \end{minipage}
   \caption{\label{ablation2} Ablation on geometry-guided deformation prediction. (Left) Quantitative results; (Right) Visual comparison.}
\end{figure*}
\begin{figure*}[h]
   \begin{minipage}{0.47\textwidth}
   \small
     \begin{tabular}{lcccc}
      \toprule
      Models & PSNR$\uparrow$ & SSIM$\uparrow$ & LPIPS$\downarrow$ & FID$\downarrow$\\
      \midrule
      NA (Full data) & \bf{23.547} & \bf{0.928} & 0.084 &{\bf 44.921} \\
      \; {w/ 5 cameras} & 23.080 &0.926&	0.082 & 52.673\\
      \; {w/ 195 frames} & 22.467 &	0.920 & \bf{0.078} &	50.55 \\
      \midrule
      \bottomrule
     \end{tabular}
   \end{minipage}
   \begin{minipage}{0.51\textwidth}
     \includegraphics[width=\linewidth]{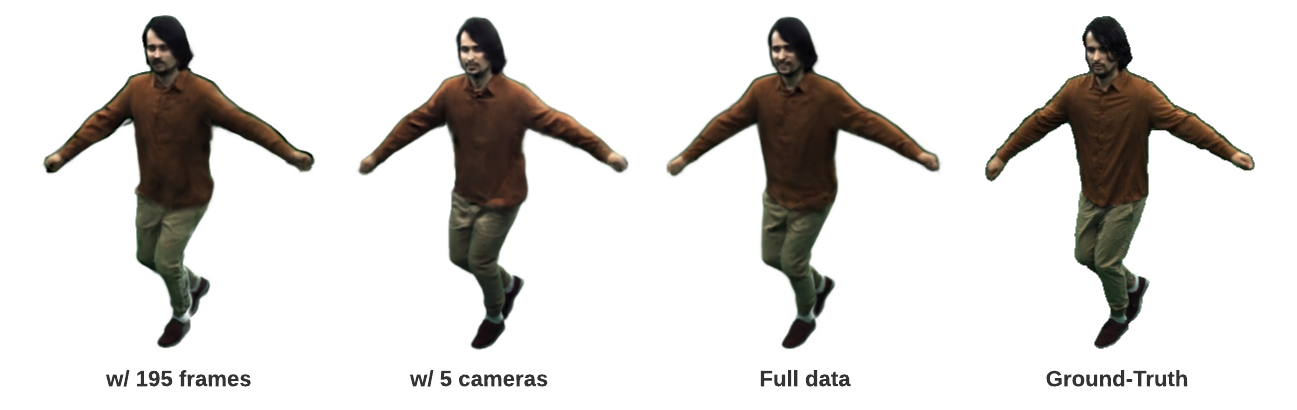} 
   \end{minipage}
   \caption{\label{ablation3} Ablation on sparse inputs. (Left) Quantitative results; (Right) Visual comparison.}
\end{figure*}
\subsection{Ablation Study}
We conducted ablation studies on \emph{D1} and evaluated on four test views for every 10th frame.
%
%
\paragraph{Effect of Texture Features.} 
In Figure~\ref{ablation1}, we first analyzed the effect of using texture features as latent variables. 
In our method, each sampled point is concatenated with the texture features extracted from the 2D texture map at its nearest surface point as conditioning for the prediction of the residual deformation and dynamic appearance.
We compare with: 1) w/o texture: neither texture nor extracted texture features are provided. 
Here, we use a pose vector as conditioning; 2) w/o feature extractor (raw texture inputs): no feature extraction is performed on the 2D texture map, that is, the texture color of the nearest surface point is used as conditioning; and 3) w/o texture w/ normal (normal map inputs): we extract the high-dimensional features on the normal map and use the features as conditioning. 
\par
We found that, compared to a compressed pose vector, the 2D texture map contains more spatial information, such as  pose-dependent local details.
Furthermore, the feature extractor can encode both local and global information and thus achieves better quality.
We also observed that using the features directly extracted from the normal map results in very poor results. 
This is because the whole normal map can represent pose information while a single pixel on the normal map does not provide any information.
%
%
\paragraph{Effect of Geometry-guided Deformation.}
We further evaluated the effect of using the SMPL model as a 3D proxy to disentangle inverse kinematic transformations and residual non-rigid deformations. 
We compare with: 1) w/o residual deformations: the spatial point in the posed space transforms to the canonical space with only an inverse kinematic transformation; 2) w/o geometry guidance: we directly predict the full movements with the deformation network.
As shown in Figure~\ref{ablation2}, modeling the full deformations as inverse kinematic transformations and the residual non-rigid deformations yields the best quality. 
Directly learning full deformations is not efficient thus results in severe artifacts. 
We further compared copying the information (skinning weights and texture features) from the nearest point on the surface with copying from the nearest vertex. Since our body model is coarse, copying information from the nearest surface points leads to an improvement. 

\paragraph{Sparse Inputs.} We tested our method with sparse training cameras and training frames as inputs. Specifically, we designed two experimental settings, one with 5 training cameras uniformly distributed on an upper sphere and the other with 195 training frames uniformly sampled in the training sequence with 19,500 frames. As shown in Figure~\ref{ablation3}, our method does not suffer from a significant performance drop with sparse inputs.

%
%
\section{Limitations}
\label{sec.limitation}
\begin{figure}[t]
    \centering
    \includegraphics[width=0.8\linewidth]{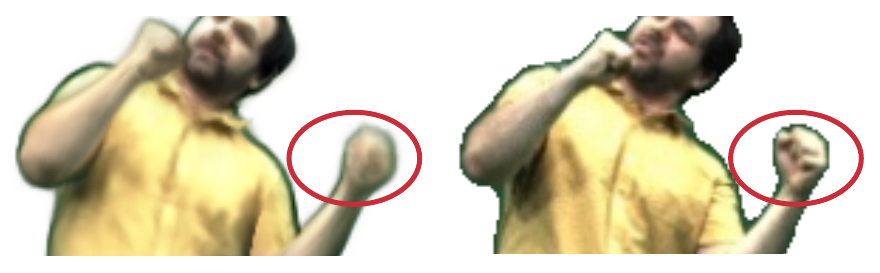}
    \caption{A failure case of rendering hands. (Left) our result; (Right) ground truth.}
    \label{fig:failure}
\end{figure}

\begin{figure}[t]
    \centering
    \includegraphics[width=0.9\linewidth]{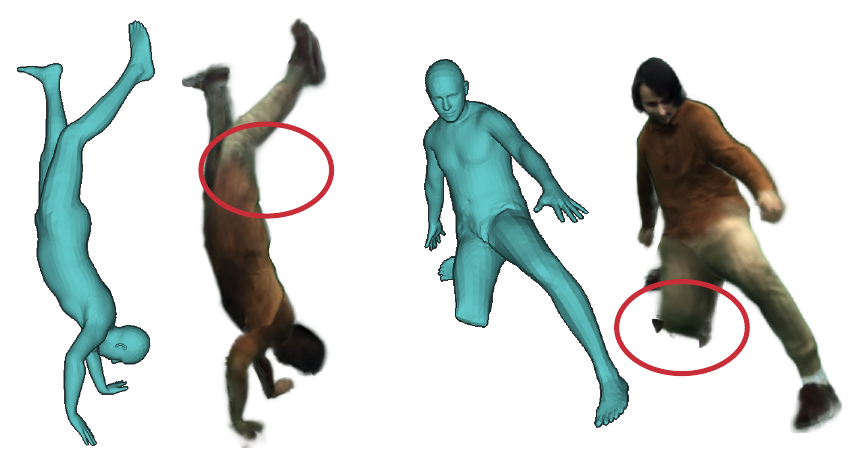}
    \caption{Failure cases on difficult poses.}
    \label{fig:failure2}
\end{figure}

Our proposed method leverages the SMPL model for unwarping to the canonical space.
Consequently, our method can handle clothing types that roughly follow the topological structure of the SMPL model, but cannot handle more loose clothing such as skirts.
Therefore future work is needed to leverage explicit cloth models on top of the SMPL model for the unwarping step.
Our method is not able to faithfully generate the fingers (see  Figure~\ref{fig:failure} for a failure example). This is because the hand is not tracked, and thus the SMPL hand is open while the GT hand is often a fist leading to severe noise in the generated GT textures for the hands, which makes the resulting input texture features to the NeRF very noisy and learning hand textures difficult.
In fact, even when using an improved human model (such as SMPL-X), robust hand synthesis can still be challenging because of the difficulty in accurately tracking hand gestrures due to the low resolution of hand images within a full body image. It will be our future work to study how to synthesize human characters with hands. 
While our method can generalize well to challenging unseen poses, like other learning-based methods, it may fail when the pose is of a totally different type from the training poses or when there is excessive joint bending, as shown in Figure~\ref{fig:failure2}. Moreover, our method is a person-specific model. Extending it to a multi-person model should be explored as future work. 
%
%
\section{Conclusion}
We presented Neural Actor, a new method for high-fidelity image synthesis of human characters from arbitrary viewpoints and under arbitrary controllable poses. 
To model moving human characters, we utilize a coarse parametric body model as a 3D proxy to unwarp the 3D space surrounding the posed body mesh into a canonical pose space. 
Then a neural radiance field in the canonical pose space is used to learn pose-induced geometric deformations as well as both pose-induced and view-induced appearance effects in the canonical space.
In addition, to synthesize high-fidelity dynamic geometry and appearance, we incorporate 2D texture maps defined on the body model as latent variables for predicting residual deformations and the dynamic appearance.  
Extensive experiments demonstrated that our method outperforms the state-of-the-arts in terms of rendering quality and produces faithful pose- and view-dependent appearance changes and wrinkle patterns. 
Furthermore, our method generalizes well to novel poses that starkly differ from the training poses, and supports the synthesis of human actors with controllable new shapes.

\begin{acks}
We thank  Oleksandr Sotnychenko, Kyaw Zaw Lin, Edgar Tretschk, Sida Peng, Shuai Qing, and Xiaowei Zhou for the help; Jiayi Wang for the voice recording; MPII IST department for the technical support. Christian Theobalt, Marc Habermann, Viktor Rudnev, and Kripasindhu Sarkar were supported by ERC Consolidator Grant 4DReply  (770784). Lingjie Liu was supported by Lise Meitner Postdoctoral Fellowship.
\end{acks}

\bibliographystyle{ACM-Reference-Format}
\bibliography{sample-bibliography}
\appendix
\newpage
\section{Additional Implementation Details}
\begin{figure*}[t]
    \centering
    \includegraphics[width=\textwidth]{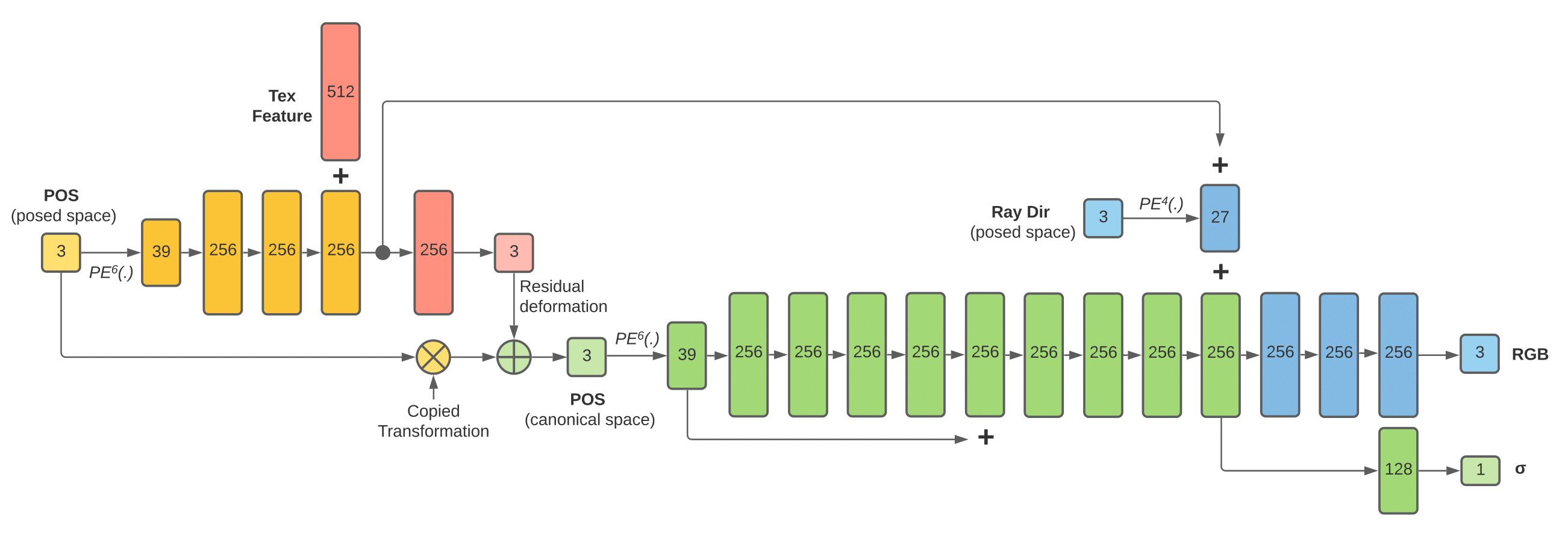}
    \caption{Visualization of the architecture of the proposed deformation and NeRF networks. 
    The number inside each block signifies the vector's dimension. All the layers are standard fully-connected layers with ReLU activation except for the output layer where we do not use activation to predict deformation, density and colors (RGB). ``$+$'' denotes vector concatenation. 
    The positional encoding function is defined as
    $\text{PE}^L(\bm{x})=\left[\bm{x}, \sin\left(2^0\bm{x}\right), \cos\left(2^0\bm{x}\right),
    \ldots,
    \sin\left(2^{L-1}\bm{x}\right), \cos\left(2^{L-1}\bm{x}\right)
    \right]$.
    }\vspace{-4pt}
    \label{fig:architecture}
\end{figure*}

\subsection{Architecture}
As shown in Figure~\ref{fig:pipeline}, our method consists of $4$ components: (1) an image translation network; (2) a texture feature extractor; (3) a deformation network; and (4) a NeRF model.  We describe each component in detail in the following.
\paragraph{Image translation network.} We adopt vid2vid~\cite{wang2018vid2vid} with the default setting using the official implementation\footnote{\url{https://github.com/NVlabs/imaginaire}}. The size of the normal map and the texture map is $512 \times 512$.
\paragraph{Texture feature extractor. } We use the feature extractor of ResNet34 backbone pretrained on ImageNet to extract features from texture maps. The extractor is jointly trained with the deformation network and the NeRF model. We extract feature maps prior to the first 4 pooling layers, upsample using bilinear interpolation and concatenate them to form the feature maps of $512$ channels. 
\paragraph{Deformation network \& NeRF. } See Figure~\ref{fig:architecture} for the network architecture for the deformation network and NeRF. 

\subsection{Algorithms}
\label{algorithm}
\paragraph{Volume rendering.}
As described in \cref{sec.deformation}, to speed up the rendering process, we adopt a geometry-guided ray marching process for volume rendering. See Algorithm~\ref{alg.ray_marching} for implementation details.
In our implementation, we set $N=64$ and $\gamma=0.06$ or $0.08$ for all sequences. 

\begin{algorithm}[h]
    \small
    \caption{\label{alg.ray_marching}\bf Geometry-guided Ray Marching}
    \KwInput{camera $\bm{p}_0$, ray direction $\bm{d}$, mesh vertices $\mathcal{V}$, $\gamma$, $N$}
    \KwIntialize{$z_{\min}=+\infty,z_{\max}=-\infty$}
    \For{$\bm{v}\in\mathcal{V}$}{
        $z_0=(\bm{v}-\bm{p}_0)^\top\cdot \bm{d}$ \\
        \If{$\|\bm{v}-\bm{p}_0\|_2^2 - z_0^2 < \gamma^2$}{
            $\Delta z=\sqrt{\gamma^2-\left(
            \|\bm{v}-\bm{p}_0\|_2^2 - z_0^2
            \right)}$\\
            \If{$z_0+\Delta z>z_{\max}$}{
                $z_{\max}=z_0+\Delta z$
            }
            \If{$z_0-\Delta z<z_{\min}$}{
                $z_{\min}=z_0-\Delta z$
            }
        }
    }
    \If{$z_{\min}<z_{\max}$}{
        Uniformly sample $N$ points in $\left[z_{\min}, z_{\max}\right]$ and perform volume rendering.
    }\Else{
        Ray missed the geometry. Abort.
    }
\end{algorithm}
\begin{algorithm}[h]
    \small
    \caption{\label{alg.find_nn}\bf Distance to Nearest Surface Point}
    \KwInput{sampled point $\bm{x}$, mesh $\{\mathcal{V},\mathcal{F}\}$}
    \KwIntialize{$l_{\min}=+\infty$}
    \For{$\bm{f}\in \mathcal{F}$}{
        $\bm{x}_0=\texttt{Project\_to\_plane}(\bm{x}, \bm{f})$\\
        \If{$\bm{x}_0$ is inside $\bm{f}$}{
            $l=\|\bm{x}-\bm{x}_0\|_2$
        }\Else{
            $\bm{a}=\mathcal{V}_{[\bm{f}_1]}$,
            $\bm{b}=\mathcal{V}_{[\bm{f}_2]}$,
            $\bm{c}=\mathcal{V}_{[\bm{f}_3]}$ \\
            $\bm{x}_c=\texttt{Project\_to\_edge}(\bm{x}, \bm{ab})$ \\
            $\bm{x}_b=\texttt{Project\_to\_edge}(\bm{x}, \bm{ac})$ \\
            $\bm{x}_a=\texttt{Project\_to\_edge}(\bm{x}, \bm{bc})$ \\
            $l=\min_{\bm{\hat{x}}\in\{\bm{x}_a,\bm{x}_b,\bm{x}_c\}}\|\bm{x}-\bm{\hat{x}}\|_2$
        }
        \If{$l<l_{\min}$}{
            $l_{\min}=l$
        }
    }
    \Return{$l_{\min}$}
\end{algorithm}
\paragraph{Finding the nearest surface point.}
For each point on the ray, we search the nearest surface  point and its $(u,v)$ coordinate from the associated SMPL model following Alogrithm~\ref{alg.find_nn} where \texttt{Project\_to\_plane} and \texttt{Project\_to\_edge} are the functions of finding the nearest points on the planes and line segments, respectively. 

We implement specialized CUDA kernels for both algorithms to achieve better efficiency.

\section{Additional Baseline Settings}
\paragraph{Neural Radiance Fields (NeRF) ~\cite{mildenhall2020nerf} + pose.} 
We extend the vanilla NeRF, which is designed for static scene rendering, to a pose-conditioned NeRF. Specifically, we concatenate the pose vector (a $72$-dimensional vector) with the positional encoding of $(x,y,z)$ for each frame. 
We use a Pytorch reimplementation of NeRF\footnote{\url{https://github.com/facebookresearch/NSVF}} and follow the default hyper-parameters. For a fair comparison, we employ the same sampling strategy used in our method for NeRF, as described in~\cref{algorithm}. We train NeRF for 300K iterations on 8 GPUs with the same batch size as our model.

\paragraph{Neural Volumes (NV)~\cite{lombardi2019neural}.}
We use the original code open-sourced by the authors\footnote{\url{https://github.com/facebookresearch/neuralvolumes}}. We use batch size of 4 per GPU and $128\times 128$ rays per image. We normalize the global translation of the scenes while keeping the rotation. All models on eight sequences were trained for 300K iterations on 4 GPUs.
Since NV requires images to encode the scene content into a latent vector, we provide the images captured by three uniformly distributed cameras to obtain the latent vector for each pose  at both  training and testing stage. 

\paragraph{Neural Body (NB)~\cite{peng2020neural}} We follow the author-provided code~\footnote{\url{https://github.com/zju3dv/neuralbody}} and run all the experiments using the default training settings.

\paragraph{Multi-View Neural Human Rendering (NHR)~\cite{wu2020multi}} We follow the author-provided code~\footnote{\url{https://github.com/wuminye/NHR}} and run all the experiments using the default training settings.

\paragraph{Real-time Deep Dynamic Characters (\emph{DDC})~\cite{habermann2021}}
Training \emph{DDC} has four stages: (1) EGNet was trained for $360{,}000$ iterations with a batch size of 40, which takes 20 hours; 
(2) The lighting was optimized with a batch size of 4, a learning rate of $0.0001$, and $30{,}000$ iterations, which takes around 7 hours;
(3) DeltaNet was trained for $360{,}000$ iterations using a batch size of 8 and a learning rate of $0.0001$ which takes 2 days.
(4) TexNet was trained with a batch size of 12 and a learning rate of $0.0001$ for $720{,}000$ iterations for 4 days.
These four networks were trained on 4 NVIDIA Quadro RTX 8000 with 48GB of memory.

\section{Additional information of training data}
\begin{figure*}[h]
    \centering
    \includegraphics[width=0.8\textwidth]{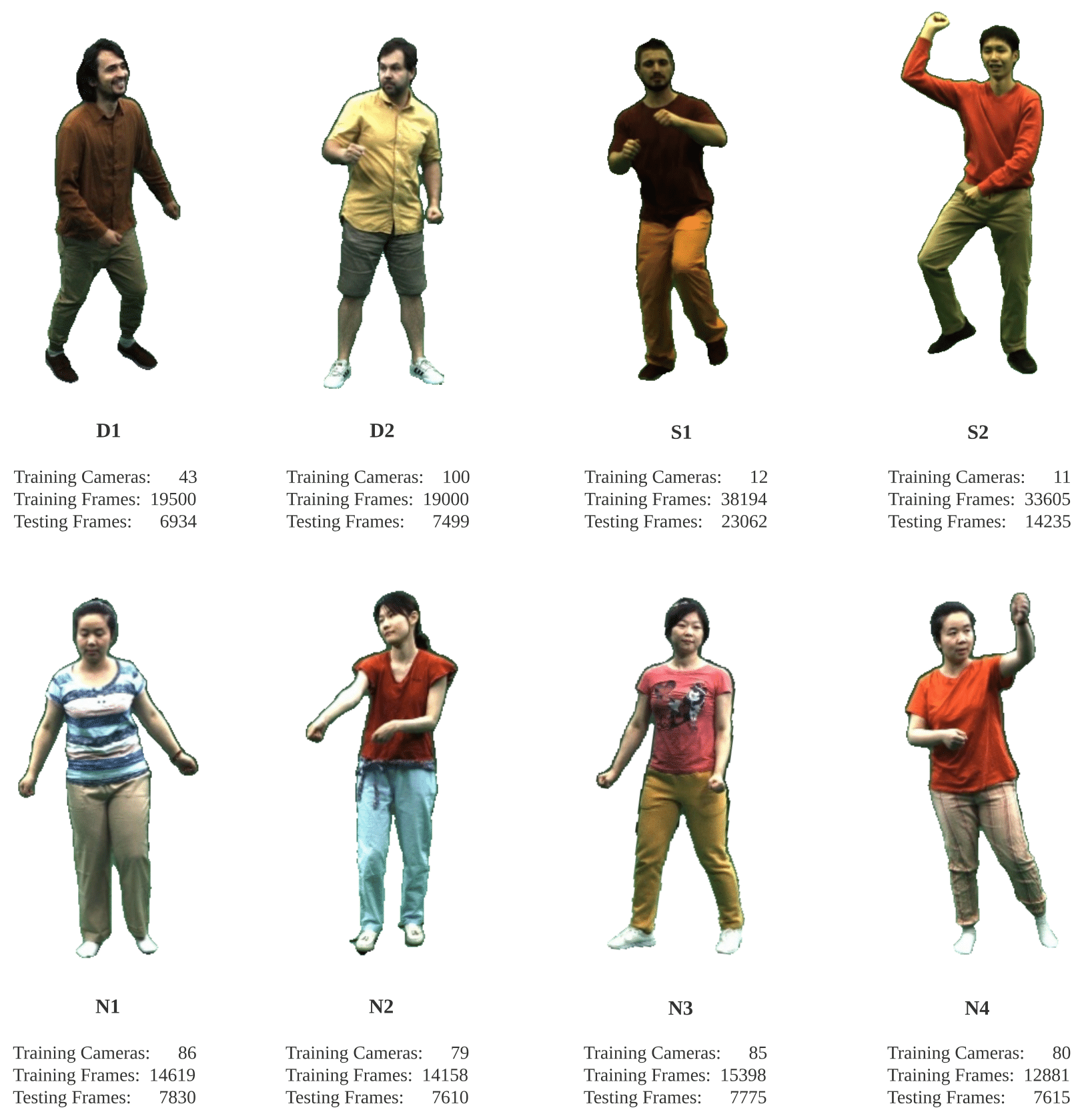}
    \caption{Detailed information of the eight sequences used in the experiments, with two sequences (\emph{D1} and \emph{D2}) from the \emph{DynaCap} dataset~\cite{habermann2021} and two sequences (\emph{S1},  \emph{S2}) from the \emph{DeepCap} dataset~\cite{deepcap} and four sequences captured by ourselves (\emph{N1-4}). For all the images, the background has been removed and the camera parameters are given. }
    \label{fig:dataset}
\end{figure*}

The detailed information of the training data is included in Figure~\ref{fig:dataset}. 

\section{Definition of Evaluation Metrics}
\paragraph{PSNR} The Peak Signal-to-Noise Ratio (PSNR) can be defined as a logarithmic quantity using the decibel scale of the mean squared error (MSE). See the  implementation at \url{https://scikit-image.org/docs/stable/api/skimage.metrics.html\#skimage.metrics.peak_signal_noise_ratio}.
\paragraph{SSIM~\cite{ssim2004}} The Structural Similarity Index Measure (SSIM) is a perception-based model that considers image degradation as perceived change in structural information, while also incorporating important perceptual phenomena, including both luminance masking and contrast masking terms. See more details at   \url{https://scikit-image.org/docs/stable/api/skimage.metrics.html\#structural-similarity}.
\paragraph{LPIPS~\cite{zhang2018perceptual}} The Learned Perceptual Image Patch Similarity (LPIPS) is a perceptual loss, which corresponds well to human visual perception. We used the implementation from \url{https://github.com/S-aiueo32/lpips-pytorch}
\paragraph{FID~\cite{FID2017}} The Frechet Inception Distance (FID) is a widely-used metric for generative models, which measures the similarity of generated images to real images. We used the implementation from \url{https://github.com/mseitzer/pytorch-fid}

\end{document}